\documentclass[lettersize,journal]{IEEEtran}
\usepackage{amsmath,amsfonts}
\usepackage{algorithm}
\usepackage{array}
\usepackage[caption=false,font=normalsize,labelfont=sf,textfont=sf]{subfig}
\usepackage{textcomp}
\usepackage{stfloats}
\usepackage{url}
\usepackage{verbatim}
\usepackage{graphicx}
\usepackage{cite}
\usepackage{makecell}
\usepackage{caption}
\usepackage{color}
\usepackage{algpseudocode}
\begin{document}

\title{From Ideal to Real: Integrating Language Models into Smart Home for Human-Centric IoT Safety and Security Fuzzing}

\author{\IEEEauthorblockN{
Lin Li,
Xinbo Ban,
Chao Chen,
Lei Pan, 
Ming Ding,
Kok-Leong Ong,
Jun Zhang,
}
\thanks{
Lin Li (Corresponding author) is with the Faculty of Science and Engineering, Southern Cross University, Gold Coast, QLD 4225, Australia (e-mail: lin.li@scu.edu.au).

Xinbo Ban is with the Research and Development Centre, The Second Research Institute
of Civil Aviation Administration of China, Chengdu 610041, China (e-mail:
banxinbo@caacsri.com)

Chao Chen and Kok-Leong Ong are with the School of Accounting, Information System and Supply Chain, RMIT University, Melbourne, VIC 3000, Australia (e-mail: chao.chen@rmit.edu.au; kok-leong.ong2@rmit.edu.au).

Lei Pan is with the School of Information Technology, Deakin University, Waurn Ponds, VIC 3216, Australia (l.pan@deakin.edu.au)

Ming Ding is with the Data61, CSIRO, Eveleigh, NSW 2015, Australia
(e-mail: ming.ding@data61.csiro.au).

Jun Zhang is with the School of Software and Electrical Engineering, Swinburne University of Technology, Melbourne, VIC 3122, Australia (junzhang@swin.edu.au)

}


}

\maketitle

\begin{abstract}
Integrating IoT devices into household systems offers enhanced control over living environments. 
However, as these devices become pervasive in critical sectors such as healthcare, transportation, and home automation, 
they expose inter-rule vulnerabilities that may lead to privacy breaches, data theft, and physical harm. 
Traditional fuzz testing methods, 
which often operate in simulated environments assuming ideal human decisions, 
fail to account for the complexities of human interactions that can influence device operations in smart homes, 
potentially leading to unforeseen consequences. 
To reveal the impact of real human decisions and discover unexpected inter-rule vulnerabilities, 
we introduce an enhanced version of IoT Fuzz for smart homes by integrating large language models to emulate human decision-making processes.  
Our novel approach adds a layer of realism and unpredictability to the testing environment, 
more accurately capturing the diverse and dynamic nature of human interactions in smart home settings.  
\end{abstract}

\begin{IEEEkeywords}
IoT, Large Language Model, Fuzzing Test, Smart Home

\end{IEEEkeywords}

\section{Introduction}
The Internet of Things (IoT) has emerged as a transformative force, redefining human interaction with everyday environments \cite{thibaud2018internet,wang2024threshold}. In smart homes, IoT devices, ranging from intelligent thermostats to automated lighting systems, work harmoniously to enhance living comfort and energy efficiency \cite{kim2021systematic}. 
Moreover, IoT in smart homes extends to security systems, where sensors and cameras provide real-time surveillance \cite{sisavath2021design}.

While smart homes bring unparalleled convenience and automation, they also present significant challenges, particularly in ensuring security and safety within these connected spaces \cite{schiller2022landscape}. As interconnected devices increase, so does the potential for vulnerabilities. Each device becomes a potential entry point for threats \cite{xiang2024blockchain}. 
The diverse range of IoT devices, including smart home appliances, wearables, and industrial sensors, often have varying security measures. 
Additionally, the interconnectivity of these devices means that a compromise in one area could lead to a domino effect, allowing attackers to move laterally within a network. 
\begin{figure}[!t]
  \centering
  \includegraphics[width=\linewidth ]{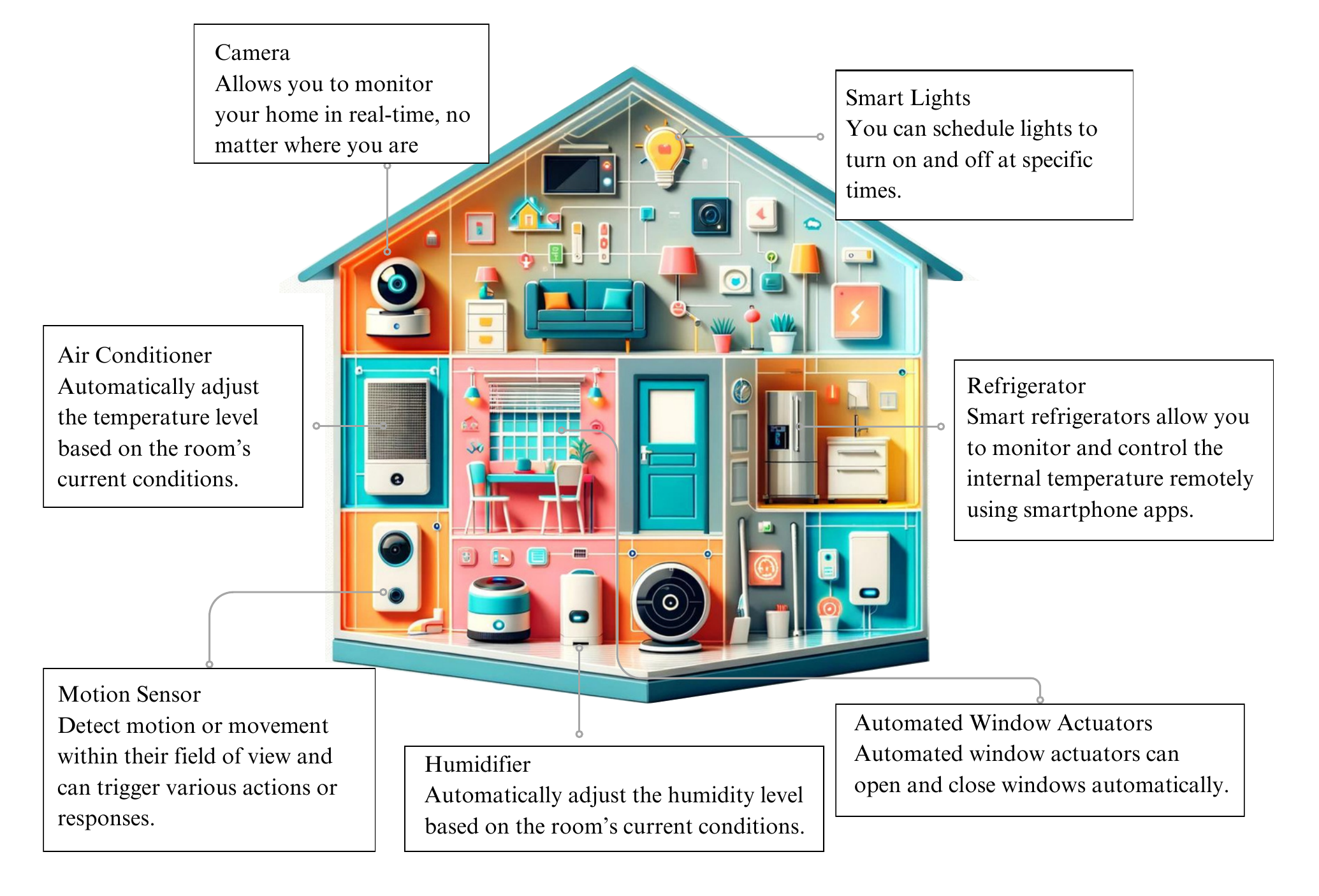}
  \caption{An example of smart home. The device can be adjusted adaptively to the indoor environment.}
  \label{smarthome}
\end{figure}

A considerable portion of existing smart home security research is centred on identifying and mitigating vulnerabilities within the hardware components of IoT devices \cite{rizvi2020identifying,wang2025privacy,long2024fedcl,wang2024trustmis}, which includes an in-depth analysis of microcontrollers, sensors, and other integral hardware elements to uncover potential weaknesses that could be exploited for unauthorized access \cite{rizvi2020threat}. 
Additionally, since smart homes usually require a network, the security of the network also forms a core part of these research, where the focus lies in examining the security of communication protocols used by smart home devices \cite{ray2020iot}.

Smart home ecosystems are governed by a set of rules and policies that dictate how devices interact and respond to various inputs \cite{leong2009rule}. These rules are often configured by the user or set by default by manufacturers.
While hardware and network security receive significant attention, the integrity and security of these governing rules are sometimes neglected. 
Vulnerabilities here can lead to scenarios where malicious actors manipulate these rules, leading to unauthorized control of the smart home system \cite{sikder2020kratos}.
For instance, some smart homes have interconnected rules where one action triggers another (e.g., "If the security camera detects motion, send an alert and record footage"). If an attacker can interfere with the initial trigger (like disabling motion detection on cameras), it can prevent the entire chain of actions, compromising security.
Different devices or applications might have conflicting rules in complex smart home systems. 

While traditional fuzzy testing is an effective technique for simulating a real environment to analyze rule vulnerabilities in smart homes \cite{shu2022iotinfer}, it typically involves random or malformed inputs, which might not always accurately replicate how real users interact with smart home systems. As a result, certain user-related vulnerabilities or issues might be overlooked. Moreover, human decisions introduce a level of unpredictability and complexity that traditional fuzzing methods may not adequately address, potentially leaving gaps in security assessments. Incorporating human interaction elements into fuzz testing can provide a more comprehensive evaluation of IoT security vulnerabilities. Thus, we are Integrating a large language model into fuzz testing to address the complexity and unpredictability of human decisions. Human interactions with IoT devices are not always predictable or rule-based. They can be random, emotional, or context-dependent. 
Large language models, trained on vast human language and behaviour datasets, can better simulate these diverse and complex interaction patterns, providing a more realistic testing environment for smart home systems. The main contributions of this paper are as follows:

\begin{itemize}
 \item We pioneer integrating a large language model into the fuzz testing framework, specifically targeting the simulation of human behaviour. This novel approach addresses the unpredictability and complexity of human decisions, which traditional fuzz testing methods may not adequately capture, thus providing a more comprehensive evaluation of IoT security vulnerabilities in smart homes.

 \item By focusing on ten specific smart home rules, we detailed insights into how different language models (Vicuna13B, Vicuna7B, and Llama2) interact with and impact these rules, helping understand the nuances of each model's interaction with smart home systems.

 \item We conducted a simulation using data across six years. This extensive simulation has provided critical insights into how these models adapt and perform over time, revealing key trends and potential security vulnerabilities. 
\end{itemize}

The rest of the paper is organized as follows: 
Related work is reviewed in Section \ref{relatedwork}. 
Section~\ref{method} describes our strategies.
In Section~\ref{experiment}, experimental results are provided.
We discuss the limitations of our method in Section~\ref{limitation}. 
Finally, Section \ref{conclusion} concludes this work.

\section{Related Work}
\label{relatedwork}

\subsection{IoT Security Analysis}
In the evolving field of smart home security analysis, researchers have developed various methodologies to address the complexities of this domain. Initially, two primary methods dominated: static analysis and runtime analysis. Static analysis, as exemplified by the use of the PScout framework and the Soot Java bytecode analysis tool, involves scrutinizing the system's design and code to unearth potential security flaws without running the system. 
In contrast, runtime analysis focuses on the system's behaviour during operation. This method, vital in detecting security issues arising in actual use, was effectively demonstrated in the analysis of the Samsung SmartThings platform. By examining 267 apps, researchers could categorize conflicts from interactions with the same device, medium, or global variables. This approach, used in tools like IotSan, employs model checking to pinpoint interaction-level flaws that could lead to unsafe states, analyzing IoT apps, sensors, and actuators to detect vulnerabilities \cite{trimananda2020understanding, nguyen2018iotsan,chen2023fedcml,wang2022towards}. However, these methods sometimes overlook direct device interactions and can't analyze closed-source devices, thus not fully capturing the complexity of an IoT environment where these factors are significant.

To address these shortcomings, researchers have developed more sophisticated tools. For instance, in the security analysis of Samsung SmartThings, traditional tools like Soot proved inadequate due to Groovy's dynamic method calling. This led to the adoption of a compiler customizer post-semantic analysis phase, enhancing the effectiveness of both static and runtime analysis by accounting for dynamic method invocation~\cite{fernandes7546527}.

Further advancements include IoTGaze, which analyzes encrypted wireless traffic to detect potential anomalies and vulnerabilities in IoT systems by identifying discrepancies in sequential interactions \cite{gu9155459}. Additionally, IoTsafe constructed a real-world physical environment to simulate intricate interactions among various IoT devices, providing a more realistic perspective of potential security risks \cite{ding2021iotsafe}. While real-world physical environments provide valuable insights for IoT security, they are not scalable for extensive analysis. Recognizing the limitations of real-world environments in scalability, the IoTfuzz \cite{ban10288143} was developed in MATLAB to create a simulated environment. This simulation facilitates the analysis of large-scale smart home security.

Despite the range of attempts by researchers to analyze smart home security, existing methods still fall short in simulating human behaviour during the analysis process. 
Human behaviour can introduce unpredictable variables, significantly affecting the security landscape of smart homes. 
Therefore, to gain a more comprehensive understanding of smart home security, our approach introduces LLMs to simulate the complexity and dynamism of human decisions.

\subsection{Human Behaviour Agent}
Human behaviour agents are advanced software systems designed to emulate, analyze, or predict human behaviours, serving purposes ranging from simulating complex human actions to aiding in research and analysis of human behaviour patterns. 
Early human behaviour agents were frequently utilized for creating non-player characters (NPCs) in video games \cite{guerrero2020studying}. These NPCs were programmed to exhibit certain behaviours and actions that mimic human players, enhancing the game's interactivity and realism. These early agents were relatively simple in their behaviour, often based on predefined scripts or basic decision-making algorithms. 
The development of research in human behaviour agents has made significant progress in recent years, particularly in areas like cognitive modelling and the creation of generative agents.
One approach to advancing human-agent interaction is through the design of cognitive models. A conceptual framework \cite{schurmann2020personalizing} has been proposed to guide researchers in this area. This framework involves evaluating the suitability of a research question for cognitive modelling, considering the resources available, and determining the relevant levels of analysis. 

Another significant advancement is in the development of generative agents. These agents are usually based on LLMs and can adapt dynamically based on their experiences and environment. LLMs are trained on extensive and diverse datasets encompassing various aspects of human interaction and communication. This training enables them to capture and replicate a wide range of human behavioural nuances, making them ideal for creating generative agents that can realistically simulate human-like responses and interactions. \cite{park2023generative} demonstrated the use of LLMs in a sandbox game environment to exhibit believable individual behaviours and emergent social interactions. 
\cite{fu2024drive} demonstrates how LLMs can interpret and interact with driving environments in a human-like manner, using common sense and reasoning. In a simulated environment, 
GPT-3.5's approach shows more human-like decision-making, evaluating consequences of actions and maintaining consistency, unlike other AI methods that tend to focus narrowly on objectives or rewards.

\begin{figure*}[!t]
  \centering
  \includegraphics[width=0.85\linewidth ]{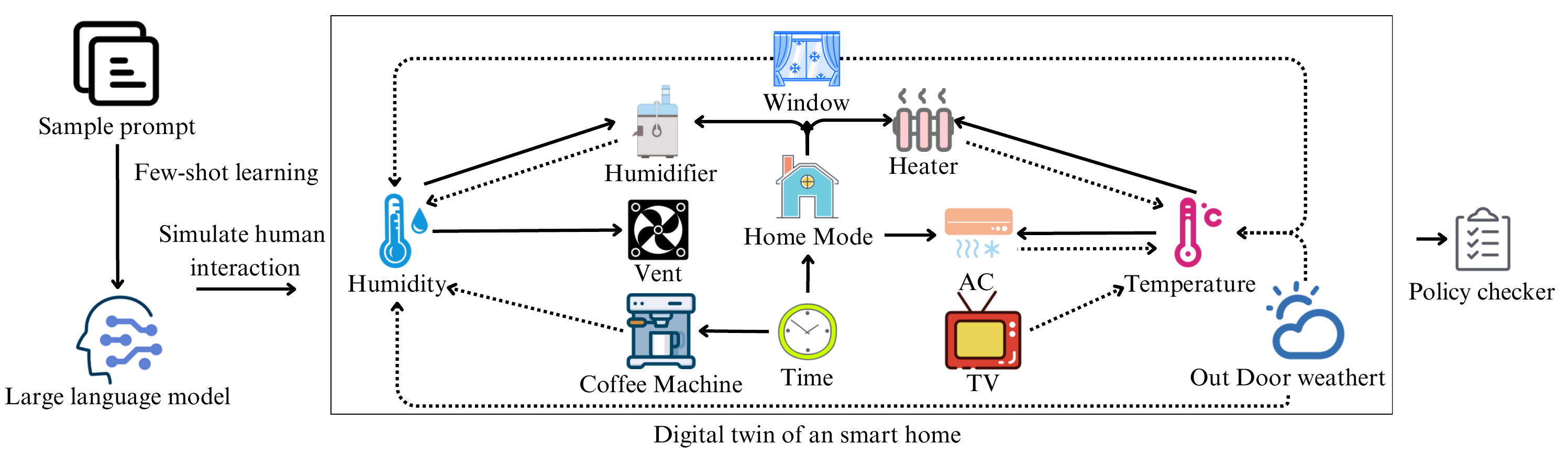}
  \caption{Workflow of our method. We generate instructions to interact with Matlab through LLM to change the state of the SH device during the simulation. The policy checker periodically assesses whether the policy is violated or not. }
  \label{Workflow}
\end{figure*}

\section{Method}
\label{method}
In this section, we provide an overview of our proposed approach. As shown in Fig.~\ref{Workflow}, our methodology comprises two key stages. Firstly, we design a series of prompts to engage a large language model, simulating human interactions within a smart home context. This approach enables us to generate diverse and realistic user commands and queries, reflecting typical user behaviour in interacting with smart home devices. Subsequently, we employ IoTFuzz \cite{ban10288143}, a tool specifically developed for simulating the operational dynamics of smart home ecosystems. IoTFuzz facilitates the emulation of various smart home devices and their interactions based on the rules set within a typical smart home environment. By integrating the outputs from the large language model with IoTFuzz, we can effectively simulate a wide range of user-device interactions, which allows us to scrutinize the smart home rules for potential vulnerabilities, identifying scenarios where conflicting or unsafe actions might be triggered by user commands.

\subsection{Agent Prompt}
To effectively simulate human interactions in the smart home context using a large language model, a crucial step in our methodology involves the meticulous design of prompts. 
In designing the LLM, we employed the method of Chain-of-Thought Prompting to achieve better results during the reasoning process \cite{wei2022chain}. This approach encourages the model to follow a step-by-step reasoning pathway that mirrors human thought processes, thus enhancing its ability to produce more accurate and contextually relevant responses. 
In our scenario, we need to make the LLM understand the current environment it is in, such as reading the temperature sensor readings and the current state of the smart home. 
For this purpose, we wrote a natural language description as input and expected to elicit responses that closely resemble human behaviour when interacting with smart home devices. 
The following is an example of an initialized conversation:

{\small\fontfamily{cmtt}
`` The time is [6:30 am]. You have just woken up in a room with the following smart devices: [Vent, Window, Humidifier, TV, Air Conditioning, Heater, Coffee Machine]. For smart devices, 1 means on and 0 means off. For windows, 0 means open, and 1 means close. For home mode, you have three states sleep, away and home. The following are the environmental states that can be detected: Indoor Temperature (range from -10 to 50), Outdoor Temperature (range from -20 to 60), Indoor Humidity (range from 0 to 100), Outdoor Humidity (range from 0 to 100). The current environment is Indoor Temperature 0, Outdoor Temperature 0, Indoor Humidity 80, and Outdoor Humidity 45. Please tell me how you will set the status of each smart device next to make the room comfortable. Please follow the format below: smart device name: (0 or 1) - your reason. Air Conditioning can lower the temperature. The vent can lower the humidifier. Please consider security as well. Turn on the heater if the temperature is below the threshold. A humidifier is turned ON when humidity is detected below the threshold. A vent is turned ON when humidity is detected to be above the threshold. The High Threshold for temperature is [32℃], low is [16℃]. The high threshold for a humidifier is [80], and the low threshold is [30]. Let's think step by step.''
}

In this prompt, the time, environment and device state are dynamically adjusted according to the running state. To further optimize the LLM's responses for our specific requirements in simulating human interactions within smart home environments, we incorporated two-shot learning. We entered the prompt into GPT 4.0 and selected two exemplary responses that best represented the ideal output we expected from the LLM in the context of smart home interactions. We then fed these selected responses back into the LLM as learning examples. This process allowed the model to analyze and learn from these ideal answers, thereby gaining a better understanding of the desired response style and content for our specific research context. Here's an example of a response used in two-shot learning:

{\small\fontfamily{cmtt}
 ``Heater: 1 - The temperature is currently 0°C, which is below the High Threshold of 16°C. Turning on the heater will raise the temperature and make the room more comfortable.  Vent: 1 - The indoor humidity is 80\%, which is above the High Threshold of 80\%. Opening the vent will release excess humidity and lower the indoor humidity to a more comfortable level. Air Conditioning: 0 - The temperature is already at the low Threshold of 16°C, so there's no need to turn on the air conditioning.
 Humidifier: 0 - The indoor humidity is already at the High Threshold of 80\%. Turning on the humidifier would add unnecessary moisture to the air.
 Window: 1 - Given that both the indoor and outdoor temperatures are 0°C, opening the window will not benefit the room's comfort level. Additionally, considering security during midnight hours, its best to keep it closed...''
}

Upon acquiring the semi-structured data, we are able to efficiently parse the status updates for each device from the outputs generated by LLMs into the simulation environment using regular expression matching.
Regular expression matching allows us to accurately extract relevant information, such as device statuses and environmental conditions, by defining precise patterns that correspond to the expected format of these updates.

\begin{table*}[!t] 
\renewcommand{\arraystretch}{1.8} 
\centering
\resizebox{\textwidth}{!}{

\begin{tabular}{|l|l|l|} 

\hline
\multicolumn{1}{|c|}{\textbf{ID}} & \multicolumn{1}{c|}{\textbf{Description}}                      & \multicolumn{1}{c|}{\textbf{MTL expression}}                                                                        \\ \hline
1                                               
&   \makecell[l]{The AC and heater must not be turned on at the same time}
& $  \makecell[l]{ \rm  \square \{\{(AC=on\}\rightarrow (Heater \neq on)\}\wedge\{(Heater=on) \rightarrow 
(AC\neq on)\}\}  } $ 
\\ \hline

2

& \makecell[l]{ When people are at home/sleep, turn on the heater if the temperature is below the threshold }
& $  \makecell[l]{ \rm \square \{ \{ (HomeMode = home)  \lor (HomeMode = sleep) \wedge \\
(Temperature < TemperatureThresholdMin) \} \rightarrow 
 (Heater = on) \} } $
                                
\\ \hline

3
& \makecell[l]{When people are at home/sleep, turn  on the AC if the temperature exceeds the threshold}                  
& $  \makecell[l]{ \rm   \square \{ \{ (HomeMode = home)  \lor (HomeMode = sleep) \wedge \\
(Temperature > TemperatureThresholdMax) \} \rightarrow 
(AC = on) \}  }  $                          
\\ \hline

4
& \makecell[l]{When people are at home/sleep, the humidifier must be ON when humidity is detected to \\ be  above the threshold}       
& $  \makecell[l]{ \rm  \square \{ \{ (HomeMode = home)  \lor (HomeMode = sleep) \wedge
\\ 
(Humidity < HumidityThresholdMin) \} \rightarrow 
(Humidifier = on) \} }   $                              

\\ \hline

5
&  \makecell[l]{When people are at home/sleep, the vent  must be ON  when humidity is detected to be \\ above the threshold}
& $  \makecell[l]{ \rm   \square \{ \{ (HomeMode = home)  \lor (HomeMode = sleep) \wedge 
\\
(Humidity > HumidityThresholdMax) \} \rightarrow 
(Vent = on) \}  }    $                                                 
\\ \hline

6
& \makecell[l]{When people are at home/sleep, the humidifier must be off if  a related room’s humidity \\  exceeds the threshold}
& $ \makecell[l]{ \rm  \square \{ \{ (HomeMode = home)  \lor (HomeMode = sleep) \wedge  
\\ 
(Humidity > HumidityThresholdMax) \} \rightarrow 
(Humidifier = off) \}  } $                                                 
\\ \hline


7
& \makecell[l]{Temperature should be within a predefined range when  people are at home/sleep}
& $ \makecell[l]{ \rm  \square \{ \{ (HomeMode = home)  \lor (HomeMode = sleep) \} \rightarrow 
\\
\{(Temperature < TemperatureThresholdMax)  \wedge \\
(Temperature >  TemperatureThresholdMin)\}\} }  $
\\ \hline

8
&  \makecell[l]{ Humidity should be within a predefined range when people are at   home/sleep}           
& $ \makecell[l]{ \rm  \square \{ \{ (HomeMode = home)  \lor (HomeMode = sleep) \} \rightarrow 
\\
\{(Humidity <  HumidityThresholdMax)  \wedge   \\
(Humidity >  HumidityThresholdMin)\}\} }  $ 

\\ \hline

9
& \makecell[l]{ The heater should be turned off when the temperature is  above a threshold, and  no one is \\ at home}
& $ \makecell[l]{\rm  \square \{ \{ (Presence = No)  \lor (Temperature > 
\\
TemperatureThresholdMax) \} \rightarrow (Heater = Off ) \}  } $

\\ \hline


10
& \makecell[l]{Window should be closed when the home mode is away}
& $\rm  \square \{ (HomeMode = Away)  \rightarrow ( Window = Closed ) \}   $                                                             
\\ \hline

\end{tabular}
}
\caption{The textual description of policies and the corresponding MTL expression.}
\label{table-policy_detail}
\end{table*}

\subsection{Digital Twin}
Based on IoTFuzz, we built a digital twin to simulate real environments. It includes human interactions, the indoor environment and actual outdoor weather conditions. Human interactions are simulated by controlling smart home devices within the indoor environment. Human presence is categorized into three distinct states: sleep, home, and away. These states are not only integral to the simulation but are also used as conditional inputs for the language model. This approach enables the model to tailor its commands based on the current human state and ensures that the simulated commands are reflective of the likely actions and choices made by humans in different states of presence or absence.
For the indoor environment in our digital twin setup, we have incorporated a total of three environmental variables alongside seven smart home devices. These environmental variables include:

\textbf{Humidity (indoor):} This is measured as a continuous value, representing indoor humidity levels. It often serves as a triggering condition for some smart home devices. 
\textbf{Temperature (indoor):} Similar to humidity, temperature is also tracked as a continuous value for indoor environments. It frequently acts as a critical factor in activating or modifying the behaviour of various smart home devices. 
\textbf{Time:} Is the third environmental variable in our model. It plays a pivotal role in the simulation, impacting device behaviour and ecological conditions throughout the day. Time-based changes affect how devices interact with each other and with their human users, considering different periods like daytime and nighttime, as well as various routines and schedules of the occupants. 
\textbf{Humidifier:} This device is designed to add moisture to the air, helping to maintain a comfortable indoor humidity level. It operates with two basic states: on and off. The Humidifier can activate or deactivate automatically upon reaching a predetermined humidity threshold, ensuring optimal indoor conditions. 
\textbf{Heater:} The Heater is crucial in maintaining the desired indoor temperature, especially in colder conditions. It can be programmed to turn on and off based on specific temperature settings or can be manually operated to suit the comfort needs of the occupants.
\textbf{Window:} Smart Windows can open and close automatically based on predefined settings or through manual control. The operation of these windows directly influences indoor temperature and humidity levels. For instance, opening the windows can lead to a decrease in indoor temperature and a potential increase in humidity, depending on the outdoor weather conditions. 
\textbf{Air Conditioner (AC):} The AC unit is essential for regulating indoor temperature. Primarily responsible for lowering the temperature. While cooling the air, the AC also tends to reduce humidity levels.
\textbf{Vent:} This system is essential for maintaining good air quality and comfort inside the home. One of its primary functions is to reduce indoor humidity levels. 
\textbf{Coffee Machine:} A smart Coffee Machine adds a level of convenience to daily routines. It can be programmed to prepare coffee at specific times or remotely activated, offering a seamless integration into the occupants' lifestyle.
\textbf{TV:} The smart TV is a central entertainment device that can be voice-controlled or integrated with other smart home systems for a seamless entertainment experience. It will slightly increase the temperature in the indoor environment.

To simulate the outdoor conditions accurately, our digital twin model incorporates historical temperature and humidity data from a specific location. These environmental variables are not static; they change dynamically over time, reflecting the actual climatic conditions of the chosen area. This approach ensures that the simulation remains grounded in real-world data, allowing us to observe and analyze how outdoor temperature and humidity fluctuations impact the indoor environment and the performance of smart home devices.

\subsection{Safety-critical Policy}

This section delineates the methodologies for vulnerability identification, employing both policy guidelines and the Metric Temporal Logic (MTL) language. We convert textual policies into MTL formulations~\cite{abbas2014robustness}, enabling analysis of event sequences and occurrences. 
MTL distinguishes itself from other temporal logics by its detailed representation of temporal and causal relationships within SHs, inclusive of timing constraints.

MTL formulations are constructed using atomic propositions (AP), propositional logic operators, and temporal operators~\cite{koymans1990specifying}. Environmental variables serve as the basis for constructing logical statements, denoted as $ p\in AP $.
The logic also incorporates operators for conjunction ($\wedge$), disjunction ($\vee$), and negation ($\neg$).
The notation ``::=” signifies a formal definition, explicating how formulas are constructed from existing symbols or terms. 
With a given time interval I, temporal operators include eventually ($\lozenge_{I}$), until ($\cup_{I} $)  always ($ \square_{I} $), and next ($\bigcirc_{I}$).

Policies are bifurcated into actions and conditions. Consider the policy ``Window should be closed when the home mode is away", where  ``Window should be closed" and `the home mode is away" respectively represent the action and the condition, translated into the MTL formula:
\begin{equation}
\square \left \{ (Presence=away)  \rightarrow  (Window=off)  \right \}
\end{equation}

The condition is defined as one that must be unsatisfied for the action to occur. To facilitate further research, we introduce an MTL template, dividing the policy into segments based on the conditions indicated by ``when" or ``if".
Thus, the policy ``$AP_{1}$ must be on when/if $AP_{2}$ and $AP_{3}$ are true" is represented as:
$\square \left \{ (AP_{2} \wedge AP_{3})  \rightarrow  AP_{1}  \right \}$. By doing this, we deal with 10 policies, as shown in Table \ref{table-policy_detail}.

\subsection{Policy Checker}
The policy checker is developed to automate the detection of policy violations within smart home environments. 
%
We use propositional scores and global scores as metrics to validate the results. 
It employs two key metrics for validating its findings: propositional scores and global scores. 
A propositional score determines whether an Access Point (AP) complies with specified policies, and a negative propositional score signifies non-compliance. 
Conversely, the global score, which is derived heuristically from the propositional scores, serves to ascertain the overall adherence to policies. 
Finally, we convert the ``always" ($ \square $) into ``not eventually" ($\neg \lozenge  $). 

Similar to the methodology employed in IoTFuzz, we modify the ``imply" ($ \rightarrow  $) into ``and" ($ \wedge $), meanwhile negating APs that are on the right side of ``imply".
For example, the MTL formula of policy 8, (seeing Table~\ref{table-policy_detail})  
$ \square\{(Presence = on)  \rightarrow \{(Humidity < ThresholdMax)  \wedge  (Humidity > ThresholdMin)\}\}   $   
is converted into 
$ \neg\lozenge  \{(Presence =on) \wedge  \{(Humidity > ThresholdMax)  \vee   (Humidity < ThresholdMin)\}\}   $   
where ``PolicyThresholdHumidityMin" and ``PolicyThresholdHumidityMax" are abbreviated as ``ThresholdMin" and ``ThresholdMax".
This transformation approach is uniformly applied to all policies, ensuring a consistent transformation across all policies.

The policy checker calculates a continuous propositional score by computing the normalized difference, e.g.,  $(Humidity - ThresholdMax) / Humidity $.
In the case of discrete evaluations, an AP is deemed to comply with the specified policies when the propositional score is positive. 
On the other hand, a negative propositional score indicates that the AP does not meet the required policy conditions. 
More precisely, the propositional score is assigned a value of 1 when an AP condition is fulfilled (for example, when $Presence = on$), indicating compliance with the policy. 
Conversely, a score of -1 is attributed when the condition is not met, signifying a policy violation.
The propositional score~(P1, P2, and P3) of policy 8 are:
\begin{align}
P_{1}& = \left\{\begin{matrix}
 1 & if    \     Presence = on \\ 
 -1& if    \   Presence \neq  on 
\end{matrix}\right. &\\
P_{2}& = (Humidity -  ThresholdMax) / Humidity &\\
P_{3}& = (ThresholdMin - Humidity) / ThresholdMin
\end{align}

The global score is computed based on the propositional score by transforming ``or", ``and", ``not" into ``max", ``min", and ``-1", respectively.
The global score for policy 8 is~$ -1*[min\{P_1, max(P_2, P_3)\}] $.
If a policy violation appears once, the global score is negative.

The probability that the entire system violates the policy is represented as follows:

\begin{equation}
P_{violation} = {N_{violation}}/{N_{checks}} 
\end{equation}
$P_{violation}$ represents the probability or ratio of policy violation. $N_{violation}$ is the number of times that a policy was violated within the observed time frame. 
$N_{checks}$ is the total number of times policies that were checked during the same time frame.

\section{Experiment}
\label{experiment}
In this section, we present a comprehensive evaluation of our method. 
We employed various smart home environments to simulate real-world IoT scenarios. 
These environments were designed to include diverse smart home devices with synchronized data collection mechanisms, capturing device behaviour. 
We used real-world weather data from 2018 to 2023 to simulate the external environment.

\subsection{Experiment Setting}
In our research, we utilized the 2023a version of MATLAB/Simulink for the creation and simulation of a digital twin representing an SH, which is renowned for its capabilities in model-based design and multi-domain simulation at a system level. 
Integration with MATLAB/Simulink and additional features were developed using Python. 
We initialize the environment settings, as shown in Table~\ref{initsetting}. 

\begin{table}[t]
\centering
\scriptsize
\renewcommand\arraystretch{1.5}

\setlength{\tabcolsep}{3mm}{\begin{tabular}{l| c}
\hline
\textbf{Environment Factors}  & \textbf{Parameters}   \\\hline
Home Mode & 0    \\ \hline
Temperature Initial  & 25   \\ \hline
Humidity Initial  & 80   \\ \hline
Temperature Min  & -10   \\ \hline
Temperature Max  & 50   \\ \hline
Humidity Min  & 0   \\ \hline
Humidity Max  & 100   \\ \hline
PolicyThresholdTemperatureMax  & 32   \\ \hline
PolicyThresholdTemperatureMin  & 16   \\ \hline
PolicyThresholdHumidityMin  & 30   \\ \hline
PolicyThresholdHumidityMax  & 80   \\ \hline
\end{tabular}}
\caption{Environment factors. The indoor environment is initialized according to the parameters and the outdoor environment is changed by the real history.}
\label{initsetting}
\end{table}

For interacting with LLMs, we utilized the API offered by FastChat \cite{zheng2023judging}. 
This choice was motivated by FastChat's robust and efficient communication framework, which enabled seamless integration with our existing systems and facilitated effective interaction with the LLMs. 
We use several popular LLMs --- Vicuna-7B \cite{vicuna2023}, Vicuna-13B and Llama2-Chat-13B \cite{touvron2023llama}.

\begin{itemize}

\item \textbf{Vicuna-13B:} Vicuna-13B is an open-source large language model developed to offer a performance comparable to advanced models like OpenAI's ChatGPT. It is trained based on the Meta LLaMA project's base model, enhanced through fine-tuning with approximately 70,000 user-shared conversations from \emph{ShareGPT.com}. 

\item \textbf{Vicuna-7B:} Similar to Vicuna-13B, it is also built by fine-tuning a LLaMA-base model using a dataset of user-shared conversations. The primary difference between Vicuna-7B and Vicuna-13B lies in their model sizes and the consequent computational requirements and capabilities. On the other hand, Vicuna-7B, with 7 billion parameters, might be somewhat less capable of language understanding and generation but is more computationally efficient and more accessible to deploy on systems with lower resource availability.

\item \textbf{Llama2-Chat-13B:} Llama2 is a family of LLMs developed and released by Meta AI. Notably, the Llama2-Chat versions are specifically fine-tuned for dialogue use cases, demonstrating a superior performance compared to many open-source chat models on various benchmarks. 
Additionally, in human evaluations, they are comparable to some popular closed-source models like ChatGPT and PaLM. 
It has been trained on a new mix of publicly available online data, including over 2 trillion tokens. 
The fine-tuning data includes publicly available instruction datasets and over one million new human-annotated examples. 
\end{itemize}

In our simulation, the result of the simulation, i.e., whether the policy is currently violated or not, is outputted every minute based on the influence of the external environment. 
Every 30 minutes, a request is sent to the LLM to obtain the interaction operation.

\begin{figure}[!h]
  \centering
  \includegraphics[width=\linewidth ]{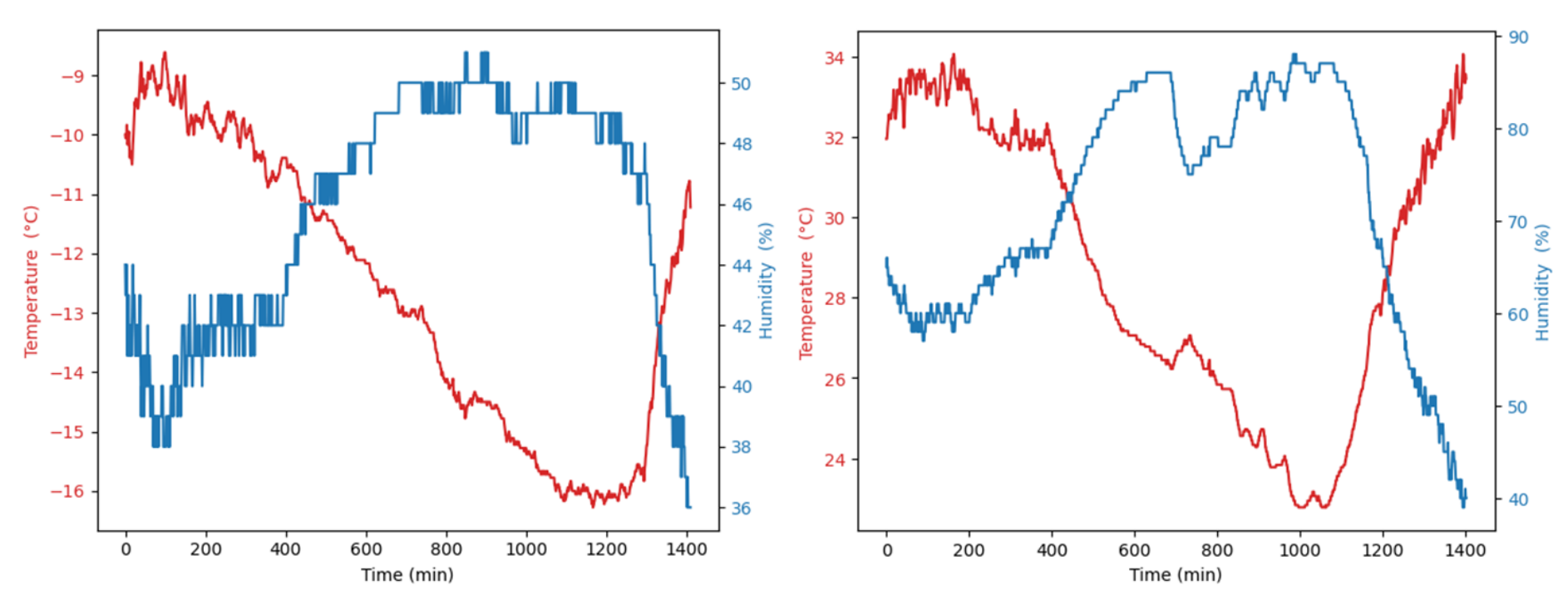}
  \caption{Examples of outdoor environmental data. The left figure is from winter, and the right figure is from summer.}
  \label{dataexample}
\end{figure}

\begin{figure*}[!t]
  \centering
  \includegraphics[width=0.9\linewidth ]{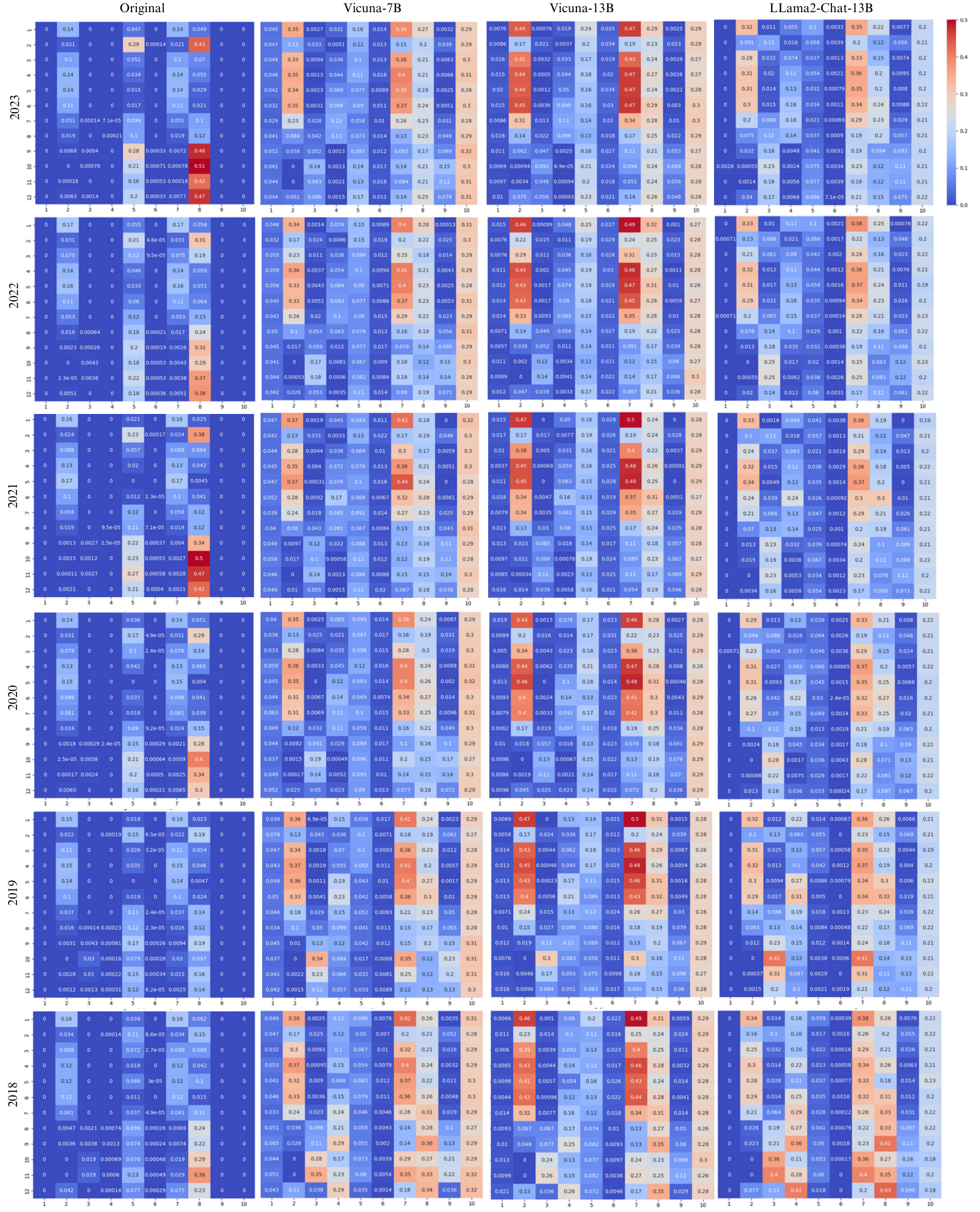}
  \caption{Simulation results from 2018 to 2023. We show the probability that each policy may be violated in each month. The $x$-axis represents the policy number. The $y$-axis represents the month. Original means that no language model is involved in the simulation.}
  \label{results}
\end{figure*}

\subsection{Outdoor Environment Dataset}
We procured an outdoor weather dataset via ThingSpeak\footnote{\url{https://thingspeak.com/channels/12397}}, utilizing the MathWorks IoT analytics platform. 
The data was retrieved using the \textbf{thingSpeakRead} function, providing a structured approach to accessing and analyzing IoT-generated weather information. 
This dataset originates from a weather station positioned atop a building in Natick, maintained by MathWorks, which consistently updates its weather readings. 
The dataset is accessible to the public and encompasses variables such as wind speed, temperature, humidity, atmospheric pressure, and rainfall. 
Measurements are recorded by the station at approximately one-minute intervals, accumulating approximately 1440 data points each day.
We extract weather data spanning from 2018 to 2023. 
Our initial step involved preprocessing this data by eliminating irrelevant columns and converting the time series data format. 
We noted variability in the daily data entry counts and some missing values, likely due to sensor malfunctions. 
Our analysis primarily concentrates on the effects of outdoor temperature and humidity. 
Thus, we have excluded data on wind speed, atmospheric pressure, and rainfall from our consideration.  
To closely simulate daily weather fluctuations and their effects, we selected files containing over 1400 data points each.

\subsection{Evaluation}
In our study, we assessed the performance of three widely used language models within our simulation framework. 
Additionally, for baseline comparison, we conducted simulations that operated without the involvement of any language model. 
As shown in Fig.~\ref{results}, we observed that the original configuration without human behaviour inputs by language models rarely violated the predefined policy, which aligns with expectations as smart home operational rules consider these policies. 
However, the integration of LLMs revealed a significant increase in strategy violations, affirming the effectiveness of our proposed method. 
This underscores the challenge of simulating human interactions within smart homes, a gap aptly filled by LLMs. 
LLMs, by learning from vast amounts of human textual behaviour, mirror the unpredictability and randomness of human actions, showcasing their utility in enhancing smart home security models by introducing realistic and complex interaction scenarios.

We also find that the probability of Vicuna-13B violating the policy is higher than that of Vicuna-7B.
The larger model, Vicuna-13B, likely has a more sophisticated understanding of contexts and can generate more complex and nuanced interactions. 
While being indicative of the model's advanced capabilities, this complexity introduces a higher risk of deviating from expected norms.
This increased complexity in its outputs could inadvertently lead to scenarios where the policy, defined by its limitations and expectations, is more easily breached. 
Interestingly, the similarity in strategy violation probabilities between LLama2-Chat-13B and Vicuna-7B, despite LLama2-Chat-13B's comparable scale to Vicuna-13B, suggests that size alone does not dictate a model's propensity to violate policy. 
This observation could imply that the architecture, training objectives, or even the data upon which LLama2-Chat-13B was trained play a significant role in determining its behaviour.

\begin{figure}[!t]
  \centering
  \includegraphics[width=\linewidth ]{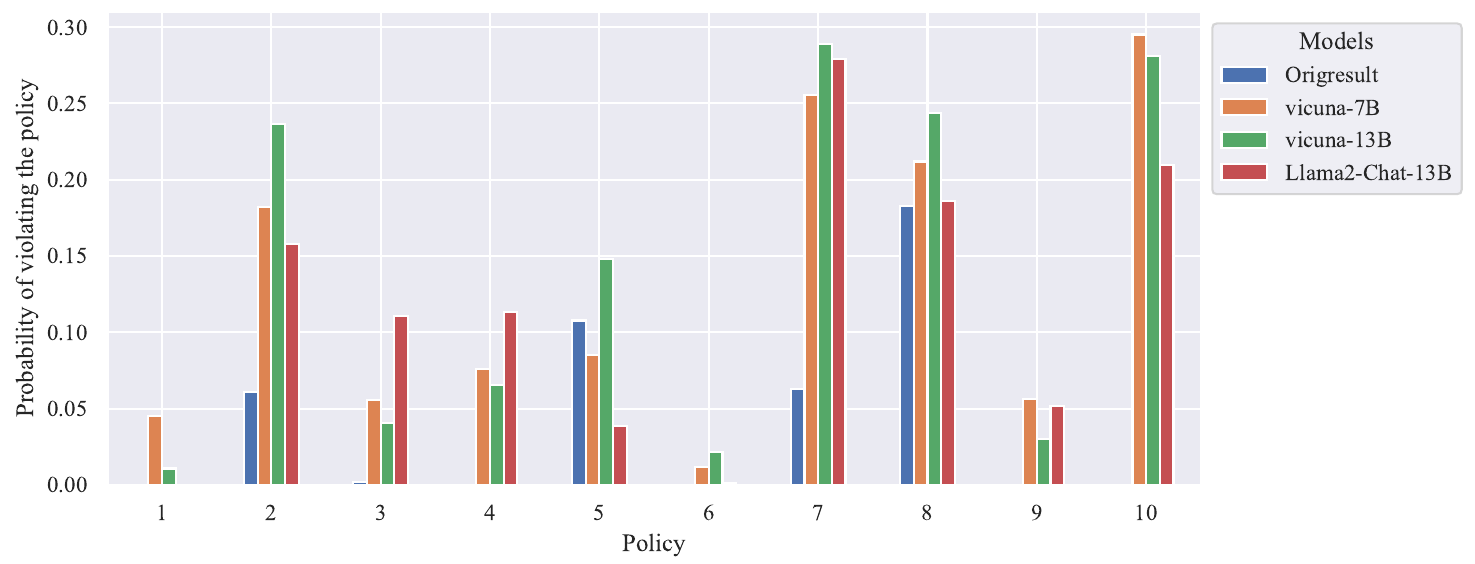}
  \caption{The overall probability that the model violates the policy. }
  \label{policyresult}
\end{figure}

In Fig.~\ref{policyresult}, we show the probability that different models violate different policies. 
The original configuration only violates policies 2, 5, 7, and 8. 
With the addition of LLM, the probability of each strategy being violated increases to varying degrees. 
Integrating LLMs into SH security assessments significantly enhances the ability to uncover potential vulnerabilities. 
By simulating complex and unpredictable human interactions, LLMs expose the intricacies and possible oversights in SH systems' rule-based operations. 
Moreover, we find that Policies 7, 8 and 10 seem more likely to be violated. This indicates that, when designing smart home systems, we can give more consideration to similar strategies.

\begin{figure}[!t]
  \centering
  \includegraphics[width=\linewidth ]{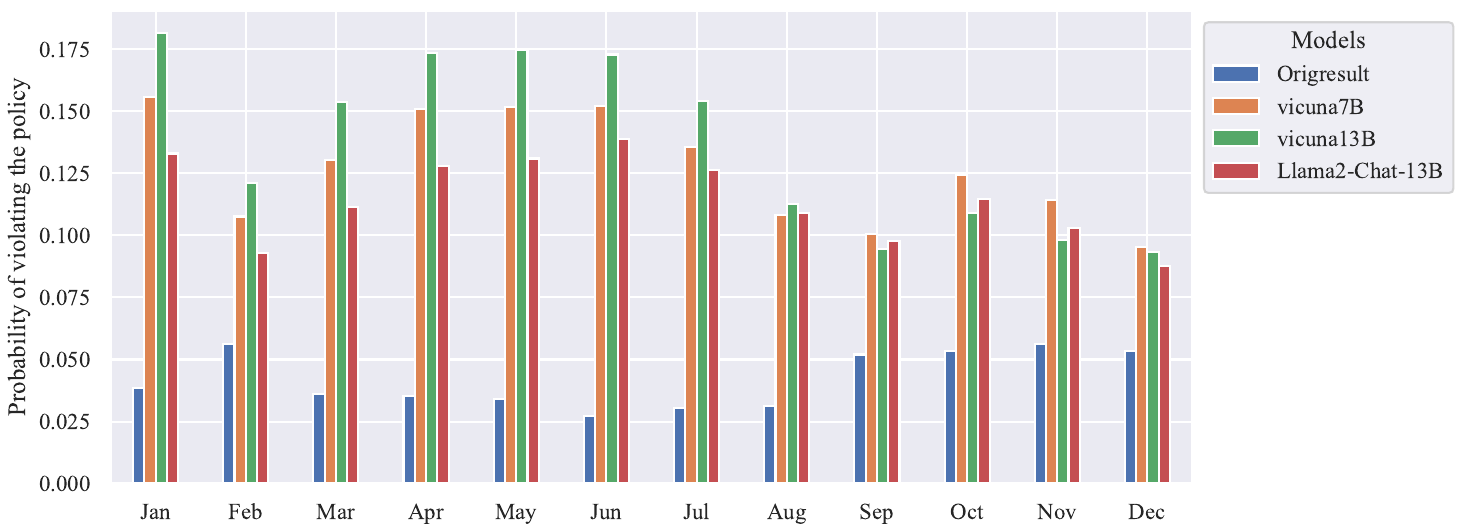}
  \caption{The overall probability that the model violates the policy. }
  \label{monresult}
\end{figure}

We demonstrate the performance of different models under different months in Fig.~\ref{monresult}. 
The original configuration of SH exhibits a higher frequency of policy violation behaviours during colder days, with fewer occurrences in summer and autumn. 
This observation aligns with the description provided in \cite{ban10288143}. 
While a well-configured SH may be secure and reliable over certain periods, it does not guarantee year-round safety. 
For LLMs, policies are more likely to be violated during hotter seasons. 
Intriguingly, we also observed a peak in policy violations during the cold month of January. 
This pattern suggests that seasonal variations in user behaviour and environmental conditions can expose vulnerabilities, indicating that SH security measures need to be adaptable to changing circumstances to maintain effectiveness throughout the year.

\section{Limitations}
\label{limitation}
Our study underscores the utility of LLMs in simulating human-like interactions within SH environments, offering an advanced methodology for uncovering security gaps. 
However, this approach is not without its limitations. 
One major constraint is the potential for hallucination from LLMs output. 
Hallucination refers to instances where these models generate text that is factually inaccurate, nonsensical, or simply not based on reality, even though it might seem plausible or coherent. 
To mitigate hallucinations, researchers have developed various techniques aimed at grounding LLM outputs in reality and improving their factual accuracy. 
These approaches include prompt engineering to guide the model towards more accurate responses, retrieval augmentation to base text generation on external evidence, feedback loops for refining responses, and novel model architectures designed to reduce the likelihood of hallucinations. 
Despite these efforts, hallucinations in LLMs cannot be completely eliminated due to the inherent limitations of current AI technologies. 
In our study, we assessed the generated responses for hallucinations by checking if we could match keywords from the responses. 
When LLMs output hallucinations, we remove them and regenerate them. 

Another limitation of our study is the environmental modelling. 
In the current simulations, we have only modelled the operation of SHs within a single room and included a limited set of smart devices. 
However, in a multi-room environment, devices may perform differently due to varying conditions across spaces. 
For instance, physical channels like illuminance might have effects that are more localized or restricted to specific areas of the home. 

Finally, to enhance the realism of environment modelling, we simulate physical models for temperature and humidity. 
We have begun to understand how real-life outdoor weather conditions affect these two physical channels. 
However, smart homes utilize a variety of other physical channels critical for their operation, such as light intensity, smoke, etc. 
Future work will aim to expand our physical modelling to include these additional channels and rooms, offering a more comprehensive and realistic simulation of smart home environments.

\section{Conclusion}
\label{conclusion}
In conclusion, we demonstrated the significant potential of integrating LLMs to augment decision support systems in smart home security. 
By simulating realistic and unpredictable human interactions, our approach effectively uncovers inter-rule vulnerabilities, offering a robust solution to the dynamic challenges in IoT security. 
To compare with traditional methods, we evaluate the system using a digital twin and a real weather environment. 
The results indicate an increased vulnerability in the smart home upon the incorporation of simulated human interactions, highlighting the critical impact of realistic behavioural dynamics on security efficacy.

\bibliographystyle{IEEEtran}
\bibliography{mybiography}

\end{document}